\documentclass{article}

\PassOptionsToPackage{numbers, compress}{natbib}

\usepackage[final]{neurips_distshift_2023}

\usepackage[utf8]{inputenc} 
\usepackage[T1]{fontenc}    
\usepackage{hyperref}       
\usepackage{url}            
\usepackage{booktabs}       
\usepackage{amsfonts}       
\usepackage{nicefrac}       
\usepackage{microtype}      
\usepackage{xcolor}         
\usepackage{amsmath}
\usepackage{appendix}
\usepackage{graphicx}
\usepackage{caption}
\usepackage{subcaption}
\usepackage{wrapfig}
\usepackage{multirow}

\title{Enhancing Robustness of\\Foundation Model Representations \\under Provenance-related Distribution Shifts}

\author{%
	Xiruo Ding$^1$\thanks{xiruod@uw.edu} \quad Zhecheng Sheng$^2$ \quad Brian Hur$^1$ \quad Feng Chen$^1$ \\
	\textbf{Serguei Pakhomov}$^2$ \quad \textbf{Trevor Cohen}$^1$\\
	$^1$University of Washington \quad $^2$University of Minnesota
}

\begin{document}

\maketitle

\begin{abstract}
  Foundation models are a current focus of attention in both industry and academia. While they have shown their capabilities in a variety of tasks, in-depth research is required to determine their robustness to distribution shift when used as a basis for supervised machine learning. This is especially important in the context of clinical data, with particular limitations related to data accessibility, lack of pretraining materials, and limited availability of high-quality annotations. In this work, we examine the stability of models based on representations from foundation models under distribution shift. We focus on \textit{confounding by provenance}, a form of distribution shift that emerges in the context of multi-institutional datasets when there are differences in source-specific language use and class distributions. Using a sampling strategy that synthetically induces varying degrees of distribution shift, we evaluate the extent to which representations from foundation models result in predictions that are inherently robust to confounding by provenance. Additionally, we examine the effectiveness of a straightforward confounding adjustment method inspired by Pearl's conception of backdoor adjustment. Results indicate that while foundation models do show some out-of-the-box robustness to confounding-by-provenance related distribution shifts, this can be considerably improved through adjustment. These findings suggest a need for deliberate adjustment of predictive models using representations from foundation models in the context of source-specific distributional differences.

\end{abstract}

\section{Introduction}
Machine learning methods have been widely applied in biomedical and clinical research. Applications have included those at the molecular level, such as predicting protein structure  \citep{jumper2021highly, akdel2022structural} and drug-drug interactions \citep{cheng2014machine}; and those at the individual level, such as electric health record (EHR) phenotyping \citep{hripcsak2013next} and supporting participant enrollment for clinical trials \citep{zhang2020deepenroll}. In the field of natural language processing (NLP), the development of deep learning methods and techniques - notably the transformer architecture introduced by \citet{vaswani2017attention} and subsequent encoder-only (BERT \citep{devlin2018bert}) and decoder-only models (GPT \citep{brown2020language}) - has advanced performance across many NLP tasks, including those in biomedicine \citep{zhang2023applications}. Most recently, generative transformer architectures pre-trained on large amounts unlabeled text, commonly referred to as Large Language Models (LLMs), have demonstrated impressive performance on many NLP tasks \citep{soltan2022alexatm, chang2023survey}. LLMs' capabilities have been demonstrated using not only traditional natural language processing (NLP) benchmarks, but also in tests of human capabilities such as the SAT and LSAT \citep{openai2023gpt4}. Foundation models are models pretrained on vast amounts of data and can be adapted for multiple downstream tasks \footnote{\url{https://crfm.stanford.edu/}}. While model weights have been released for some foundation models, such as Llama \citep{llama1} (and its successor Llama-2 \citep{llama2}) and Bloom \citep{scao2022bloom}, end-to-end fine-tuning of these models requires computational resources beyond those in many academic and clinical settings.

Contrary to the large amount of pretraining materials collected from a variety of sources for LLMs, in biomedical research high quality annotated datasets are often very limited in size and diversity \citep{sheller2020federated}. Consequently, researchers may choose to integrate data from multiple institutions. This practice can increase dataset size but also introduces a potential bias when both language use and class distribution differ across these institutions. We refer to this form of distribution shift as \textit{confounding by provenance} \citep{ding2023backdoor}. The main concern is that of site-specific label distribution shift from training to testing/deployment time. For example, if one institution has a much higher proportion of positive examples at training time, but a much lower proportion at test time, the model may make erroneous positive predictions based on language use at this institution that is unrelated to the outcome of interest.  Representations derived from LLMs encode linguistic information from outside the context of a labeled training set, and it is possible that this information may confer a degree of robustness to confounding by provenance, making a resulting model less sensitive to institution-specific linguistic differences. The current work is motivated by a desire to assess the extent of this robustness, if it is indeed conferred.

In this work, we first propose an evaluation framework for confounding by provenance. We use a BERT \citep{devlin2018bert} variant (Sentence-BERT \citep{reimers2019sentence}) and Llama \citep{llama1, llama2} models in our experiments, two widely-used and publicly-available foundation models that exemplify the encoder-only and decoder-only approaches, respectively. We test their stability under different degrees of distribution shift within the framework, across a range of synthetically-induced shifts in provenance-specific class distribution. To preserve computational resources, we extract the contextual embeddings generated from these foundation models and test them under a regression framework, with and without a simple adjustment. This procedure involves extending a method of confounding adjustment originally developed for discrete representations \citep{landeiro2016robust}, to representations from foundation models, and we assess its utility as a means to enhance their robustness to confounding shift. 

\section{Preliminaries and Methods}
\paragraph{Evaluation Framework} In this work, we focus on one specific form of distribution shift, \textit{confounding shift} \cite{landeiro2016robust}, where label distributions among subpopulations differ in the training and testing set for a text classification problems: $	P_{train}(Y|Z) \neq P_{test}(Y|Z)$, where $Y$ is the label and $Z$ is the provenance variable. Note that this problem formulation does not include the distribution of predictor variables, $X$, which in our case is derived from language.

We build upon the approach for synthetic injection of confounding shift \citep{landeiro2016robust, landeiro2018robust}, to develop an evaluation framework for binary classification with two subpopulations, assuming $Y \in \{0,1\}$ and $Z \in \{0,1\}$. The following parameters to construct a testing scenario are set:
\begin{align}
	& P_{train}(y=1|z=0) \label{eq:trainY1Z0}\\
	& P_{train}(y=1|z=1)  \label{eq:trainY1Z1}\\
	& P_{train}(z=1) = P_{test}(z=1) = C_z \label{eq:Cz} \\
	& P_{train}(y=1) = P_{test}(y=1) = C_y \label{eq:Cy} 
\end{align}
where Equation~(\ref{eq:Cz}) aims to eliminate a potential confounding factor where the proportion of training examples (irrespective of their class label) from each source is different at training and testing time, and Equation~(\ref{eq:Cy}) is implicitly enforced to negate effects of different background positive rates in the train and test sets. The objective of these constraints is to focus on shifts related to provenance. In contrast to the work of \citet{landeiro2016robust, landeiro2018robust} where a relative difference of subtraction was used, we introduce two auxiliary variables for measuring differences in site-specific class distribution:
\begin{align}
		 \alpha_{train} = \frac{P_{train}(y=1|z=1)}{P_{train}(y=1|z=0)} , \; \alpha_{test} = \frac{P_{test}(y=1|z=1)}{P_{test}(y=1|z=0)} \label{eq:alphaTest}
\end{align}
During evaluation, we specify desired ranges for variables (\ref{eq:trainY1Z0}), (\ref{eq:trainY1Z1}), (\ref{eq:Cz}), and $\alpha_{test}$. All combinations of these parameters are applied to govern selection of corresponding samples to construct multiple train/test set combinations with different degrees of confounding shift. Ultimately, the goal is to examine a model's robustness to these different degrees of distribution shift (measured by the difference between $\alpha_{train}$ and $\alpha_{test}$). To quantify robustness (or model stability), $\alpha_{test}$ is first log-transformed and a linear regression line is fit against a target evaluation metric (AUPRC value, Area Under the Precision-Recall Curve, in our case), inspired by \citet{taori2020measuring}. This coefficient measures the slope of a line that relates changes in the performance metric of interest to changes in $\alpha_{test}$. The lower the absolute value of the fitted coefficient, the more robust a model is to confounding shift, with a value of zero indicating equivalent performance irrespective of this shift.

\paragraph{Backdoor Adjustment} Originally proposed by \citet{pearl2009causality} (\textit{Causality}, Equation 3.19), backdoor adjustment is a technique to make adjustments on predictions when confounding variable ($z$) exists:
\begin{align}
	P(y|x) = \sum_z P(y|x, z) P(z) \label{eq:BA}
\end{align}

A similar approach was developed by \citet{landeiro2016robust} for text classification in the presence of confounding bias. Specifically, a logistic regression model is fit to estimate $P(y|X,z=c)$:
\begin{align}
	logit(y_c) = \beta_0 + \boldsymbol{\beta_1 X} +\boldsymbol{ \beta_2 z_c} + \epsilon \label{eq:lr}
\end{align}
where $\boldsymbol{z_c}$ is a one-hot matrix where the membership in a specific $Z$ class is represented by a value $v$, a hyperparameter that controls the degree of adjustmnet \citep{sutton2006reducing, landeiro2018robust}. Estimates of $P(y_c|x,z_c)$ from Formula~\ref{eq:lr} can then be used in Formula~\ref{eq:BA} to get an adjusted estimate for $P(y|x)$.

\paragraph{Embeddings}  We use a simple distributional language model, \textbf{Binary Unigram}, as our baseline for representing natural language, as in previous work \citep{landeiro2016robust} for point of comparison. Additionally, we use the \textbf{Sentence-BERT} model, which has been optimized for generating semantically meaningful sentence embeddings \citep{reimers2019sentence}, as a moderately-sized foundation model.


\paragraph{Llama Embeddings}  Three versions of the model are public, marked by the number of parameters as Llama 2-7B, Llama 2-13B, and Llama 2-70B. Our extension to Llama model beyond its generative ability is to extract embeddings, and then apply a classification head to them. We use the average embedding across all tokens to represent a unit of text. To investigate a potential relationship between robustness and model size, we derive representations from all three versions of Llama 2. 

For consistency with prior work \cite{landeiro2016robust, landeiro2018robust}, we use logistic regression as the classifier for all models, with and without applying backdoor adjustment.

\section{Experiments}

\paragraph{Datasets} SHAC is a dataset of electronic health record notes annotated for social determinants of health (SDoH) \citep{lybarger2021annotating, lybarger20232022}. The notes were collected from two different sources: clinical notes of chronic pain patients from the University of Washington Medical Center, and discharge notes of intensive care unit patients from MIMIC-III \citep{johnson2016mimic}. Our goal for this work is to classify patient's social history sections for whether they show any sign of drug abuse. For Hate Speech Detection, we selected two publicly available hate speech detection datasets: (1) a dataset of hate speech entries generated through perturbation on publicly available datasets \citep{vidgen2020directions, vidgen-etal-2021-learning}; (2) a dataset collected from a random set of posts extracted from a white supremacist forum, and labeled  for hate speech \citep{de-gibert-etal-2018-hate}.
 
%

\paragraph{Simulations} Using the evaluation framework, we simulated different degrees of provenance-related distribution shift. Specifically, we set value ranges for $P_{train}(y=1|z=0), P_{train}(y=1|z=1), P_{train}(z=1), \alpha_{train}$.  For each simulated setting, an experiment was repeated five times for robustness of results. 
In addition, we varied $C_y$, the overall prevalance of the positive class irrespective of provenance (more positive examples lead to better performance). Three $C_y$ values were selected to represent different background prevalence rates: 0.2, 0.48, and 0.72 for evaluations in the SHAC dataset, and 0.36, 0.44, 0.52 for Hate Speech Detection dataset. For further details of simulation configurations, refer to Appendix~\ref{appendix:simConfig}.

\paragraph{Results and Discussion}
\begin{figure}[h]
	\centering
	\begin{subfigure}[t]{.8\textwidth}
		\includegraphics[width=\textwidth]{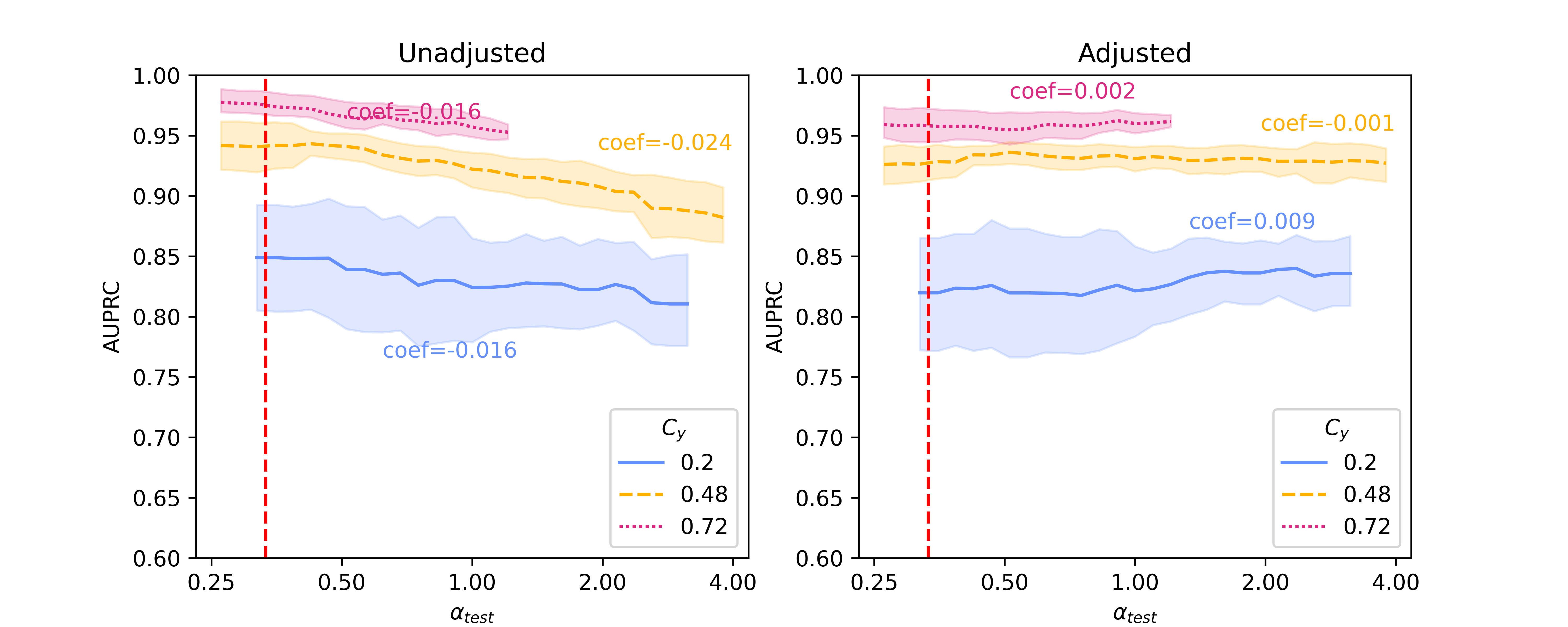}
		\subcaption{Binary Unigram}
		\label{afig:SHAC_binary}
	\end{subfigure}
	\begin{subfigure}[t]{.8\textwidth}
		\includegraphics[width=\textwidth]{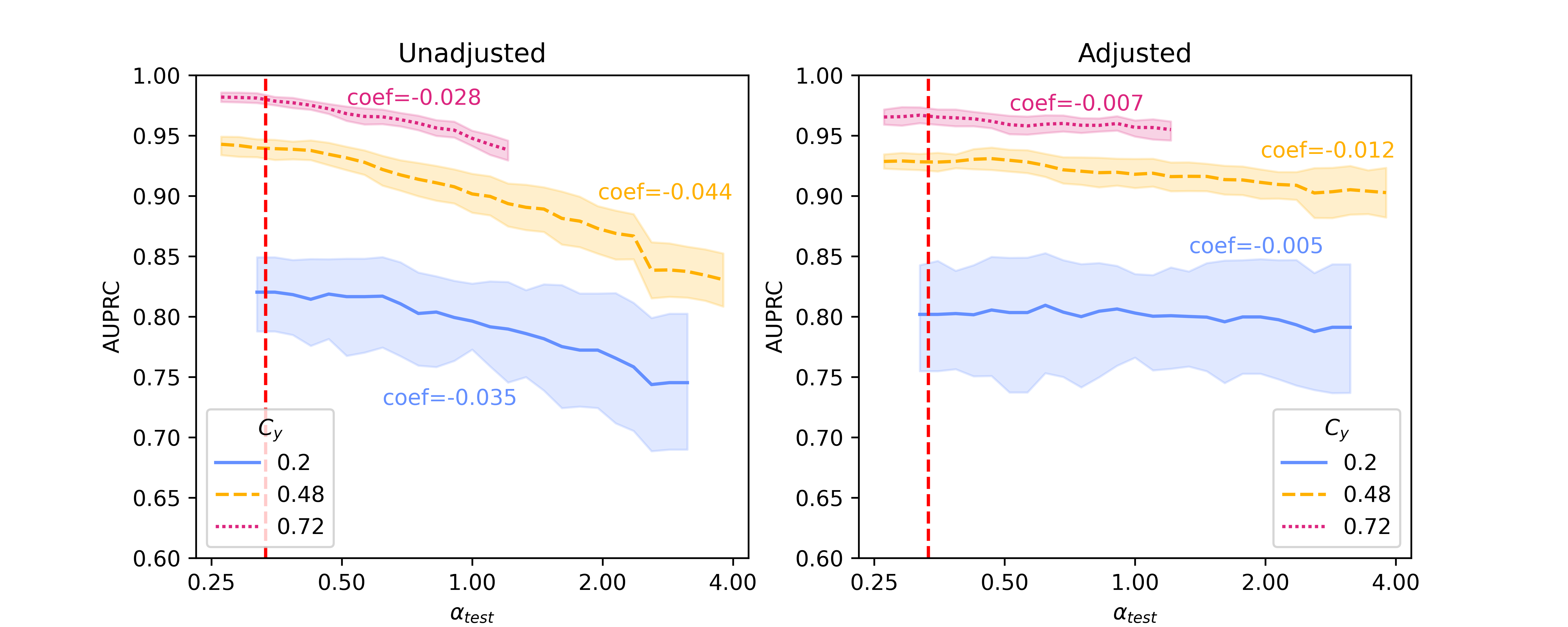}
		\subcaption{Sentence-BERT}
		\label{fig:SHACLR_sentenceBERT}
	\end{subfigure}
	\begin{subfigure}[t]{.8\textwidth}
		\includegraphics[width=\textwidth]{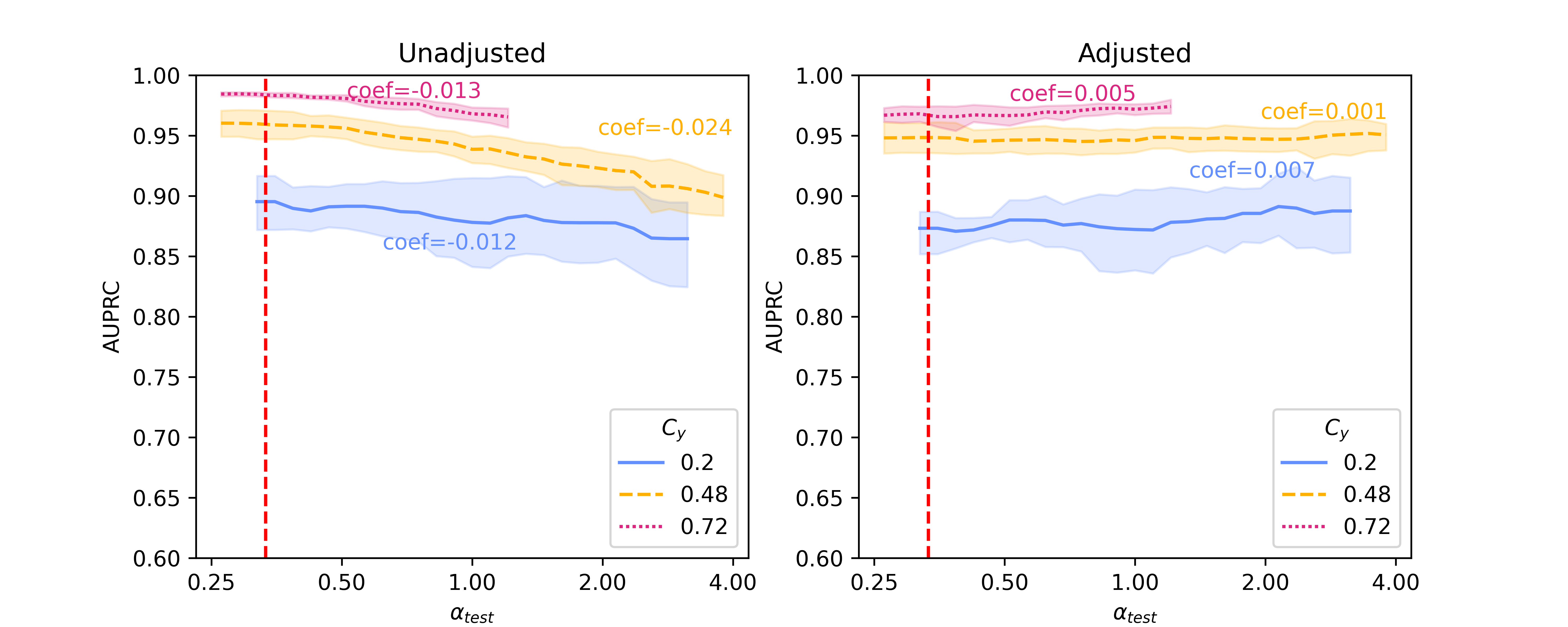}
		\subcaption{Llama 2-7B Average}
		\label{fig:SHACLR_llama}
	\end{subfigure}
	\caption{Logistic regression and Backdoor Adjustment on SHAC. (a) Binary Unigrams; (b) Embeddings generated from Sentence-BERT; (c) Embeddings generated from Llama~2-7B, then averaged. Red vertical line indicates $\alpha_{train}=0.33$, in which setting the model is trained.}
	\label{fig:SHACLR}
\end{figure}

Figure~\ref{fig:SHACLR} shows results of the unadjusted (left column) and adjusted (right column) logistic regression models using Binary Unigram, Sentence-BERT, and Llama~2 embeddings 
when $\alpha_{train}=0.33$. This $\alpha_{train}$ value indicates that we are shifting toward greater test set representation of positive examples from the second site ($z=1$) as we move to the right along the x-axis from the dashed red line. An approximation of a mirror image of this figure can also be generated by setting $\alpha_{train}=3$ (more positive examples from the first site 
at training time) and moving to the left to decrease representation of positive examples from this site ($z=1$) (Figure~\ref{afig:SHACLR_a3} and Figure~\ref{afig:HSLR_a3}). With both Sentence-BERT and Llama, the unadjusted regression models (leftmost panels) decrease in performance as the provenance-specific class distribution moves toward over-representation of the second site ($z=1$) relative to $\alpha_{train}$, resulting in moderate negative slopes. However, the absolute values of the coefficients of regression lines fit to the performance when using Llama~2 embeddings is lower than those from Sentence-BERT (e.g., 0.013 vs 0.028 for $C_y=0.72$, 0.024 vs 0.044 for $C_y=0.48$
), suggesting that the Llama2 embeddings may be innately more robust to confounding shift than those from Sentence-BERT. However, in comparison with baseline binary unigrams, the absolute values of the coefficients from Llama 2-7B embeddings are only slightly smaller, e.g., 0.012 vs 0.016 for $C_y=0.2$, 0.013 vs 0.016 for $C_y=0.72$, suggesting that representations from foundation models have no innate advantage in robustness to confounding shift as compared with unigram representations. Results from the Hate Speech Detection dataset (Figure~\ref{fig:HSLR}) show a similar trend, with the difference that those generated using foundation models have slightly smaller absolute coefficients than the unigram baseline.

With Backdoor Adjustment applied (rightmost panels), the line fit to performance is flattened, and the absolute values of the coefficients decrease (e.g. dropping from 0.035 to 0.005 for $C_y=0.2$ using Sentence-BERT embeddings), demonstrating a marked increase in robustness to confounding shift. This technique also increases the robustness of models trained on Llama~2 embeddings (shown in Figure~\ref{fig:SHACLR_llama} right pane), though the difference between models with and without adjustment is relatively small. When comparing with baseline binary unigrams, absolute coefficients on results from both Sentence-BERT and Llama~2 embeddings do not drop, and increase in some scenarios. Results on the Hate Speech Detection dataset (Figure~\ref{fig:HSLR}) show a similar trend. However, with respect to performance (rather than robustness) models using Sentence-BERT and Llama~2 embeddings both comfortably outperform baseline binary unigrams, measured directly by AUPRC (the height of AUPRC lines in general). 

\section{Conclusion}
In this work, we investigate robustness of foundation models, from Sentence-BERT to different versions of Llama~2, for the task of text classification under a framework for provenance-related confounding shift. From empirical experiments on one biomedical and one general domain dataset, embeddings from foundation models show some out-of-the-box robustness to confounding shifts, demonstrated by their smaller absolute coefficients when no adjustment is made. Additionally, they increased baseline performances on one of our datasets. Employing Backdoor Adjustment within a logistic regression framework further enhances this robustness for both foundation models and baseline binary unigram models, although the latter gain more from this adjustment. It's hypothesized that the sensitivity of the contextual embeddings from Sentence-BERT and LLAMA might impede the accurate modeling of robustness when employing a logistic regression classifier. Future work will explore alternative classifiers and methods for adjusting foundation models.

\begin{ack}
This work was supported by U.S. National Library of Medicine Grant (R01LM014056).

\end{ack}

\newpage
\bibliography{ref}


\clearpage
\appendix

\setcounter{table}{0}
\renewcommand{\thetable}{A\arabic{table}}
\setcounter{figure}{0}
\renewcommand{\thefigure}{A\arabic{figure}}

{\Large \bf Appendix}

\section{Simulation Configurations} \label{appendix:simConfig}
In the simulation of different degrees of distribution shifts by provenance, 4 parameters are required to set up the framework: $P_{train}(y=1|z=0), P_{train}(y=1|z=1), P_{train}(z=1), \alpha_{test}$. The first three were sampled from 0 to 1 evenly in linear space, with a step size of 0.05 or 0.1, depending on scenarios. 

\begin{figure}[h]
	\centering
	\includegraphics[width=.5\textwidth]{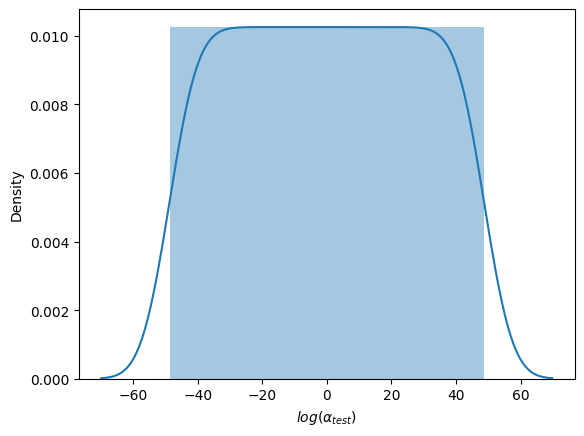}
	\caption{$\alpha_{test}$ distribution.}
	\label{fig:alphaTestDistribution}
\end{figure}

Since $\alpha$ represents the ratio for positive rates from two sources and $\alpha=1$ means same prevalence rates, we sampled $\alpha_{test}$ in a reciprocal  manner while centering around 1.0. One such example is $\{..., 1/8, 1/4, 1/2, 1, 2, 4, 8, ...\}$. This can ensure a uniform distribution of $\alpha_{test}$ in the log space, as shown in Figure~\ref{fig:alphaTestDistribution}.

\begin{figure}[h]
	\centering
	\includegraphics[width=.8\textwidth]{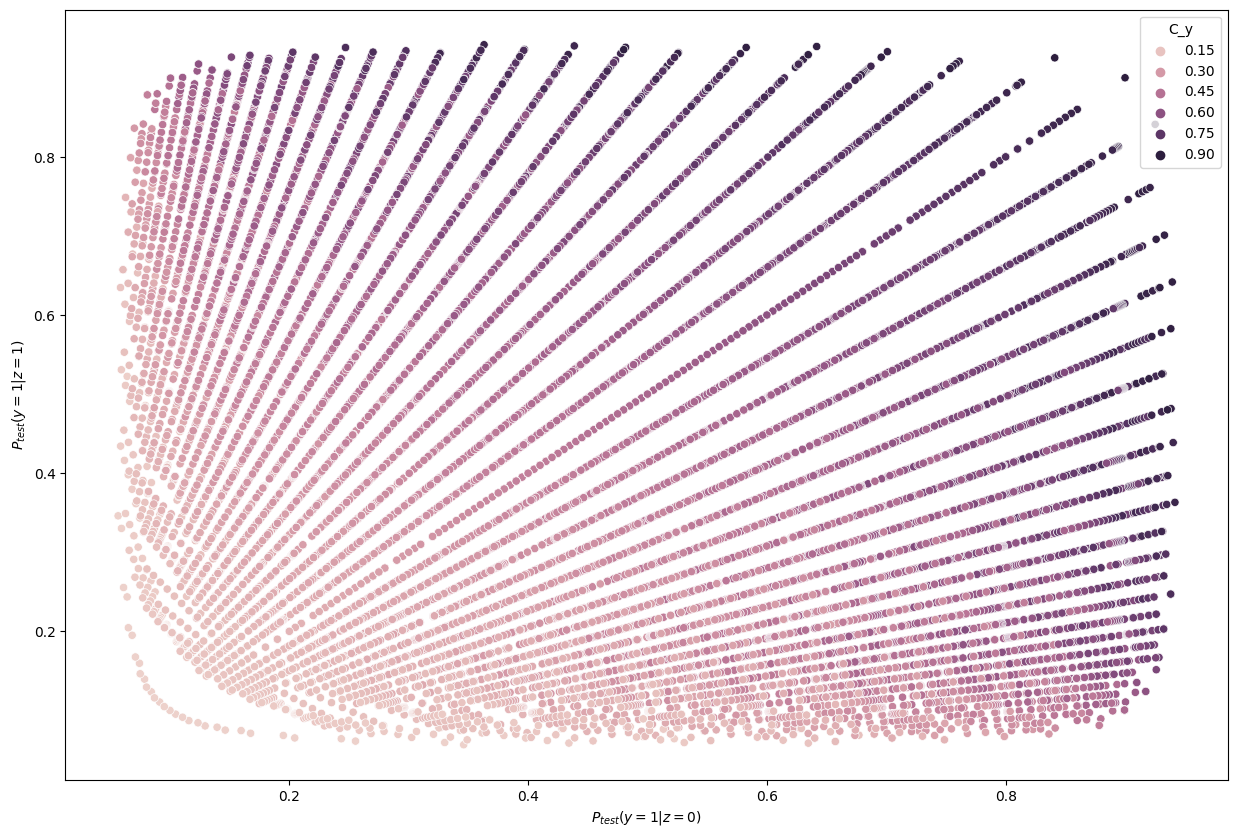}
	\caption{Theoretical sampling from the joint distribution of $P_{test}(y=1|z=0)$ and \mbox{$P_{test}(y=1|z=1)$.}}
	\label{fig:theoreticalSampling}
\end{figure}

To break down $\alpha_{test}$ into detailed samplings of its two subpopulations, Figure~\ref{fig:theoreticalSampling} shows the theoretical sampling for the joint distribution of $P_{test}(y=1|z=0)$ and $P_{test}(y=1|z=1)$. It is worth noting that not all sampling configurations from Figure~\ref{fig:theoreticalSampling} can lead to a valid training/testing set, given how many positive/negative samples from both sources required for that setting.

\section{Details on Datasets} \label{appendix:dataset}
\paragraph{SHAC} SHAC is a dataset of electronic health record notes annotated for social determinants of health (SDoH) that provided a basis for the recent n2c2/UW SDoH Challenge \citep{lybarger2021annotating, lybarger20232022}. The Social History Annotation Corpus (SHAC) was designed for extracting Social Determinants of Health (SDOH) in clinical notes under an active learning framework. The notes are from two different sources: clinical notes of chronic pain patients from the University of Washington Medical Center (UW set), discharge notes of intensive care unit patients from MIMIC-III (MIMIC set). Its annotation guideline includes several event types: (1) substance use (alcohol, drug, tobacco); (2) employment; (3) living status. Among those, we selected a small section, Drug Abuse, as the classification target in this work. 
Summaries are shown in Table~\ref{stab:dataSHAC}. To ensure enough samples for most testing scenarios, we set the training set size to 800 and testing set size to 200 for the SHAC dataset.

\begin{table}[h]
	\caption{SHAC dataset summary}
	\label{stab:dataSHAC}
	\centering
	\begin{tabular}{lrrr}
		\toprule
		& Total Number & Identified Drug Abuse Cases & Positive Rate \\
		\midrule
		UW    & 2,528        & 1,040                       & 41.1\%        \\
		MIMIC & 1,877        & 371                         & 19.8\%       \\
		\bottomrule
	\end{tabular}
\end{table}

\paragraph{Hate Speech} We collected two publicly available datasets for hate speech detection. The first one (DynGEN set) is a dynamically generated dataset by \citet{vidgen-etal-2021-learning}, through four rounds of data creation. The first round collected synthetic texts, created by the annotation team to closely mimic real-world content. It was then followed by perturbations in the texts to create new examples for next three rounds. In the end, it results in a total of around 40,000 entries, with a positive rate of 45\%.

The second dataset (WSF set) comes from a real-world white supremacist forum published between 2002 and 2017 \citep{de-gibert-etal-2018-hate}. Posts were collected from a subset of 22 sub-forums covering diverse topics and nationalities, segmented into sentences, and then manually labeled. Authors used 4 types for annotation: (1) \textsc{Hate}; (2) \textsc{NoHate}; (3) \textsc{Relation}, where the sentence itself does not convey any information and must be put in its context to be correctly identified, such as a reply to a hate speech comment; (4) \textsc{Skip}, where \textsc{Hate}/\textsc{NoHate} cannot be decided. To reduce noise, we only keep \textsc{Hate} and \textsc{NoHate} label for our work, which is the majority of the texts. \textsc{Relation} has 168 records and \textsc{Skip} has 73.

Table~\ref{stab:dataHS} provides the summary of Hate Speech Detection dataset. For this dataset, we set training set size to 4000 and testing set size to 1000.

\begin{table}[h]
	\caption{Hate Speech dataset summary}
	\label{stab:dataHS}
	\centering
	\begin{tabular}{lrrr}
		\toprule
		& Total Number & Identified Hate Speech & Positive Rate \\
		\midrule
		DynGEN    & 41,144        & 18,969                       & 46.1\%        \\
		WSF & 10,703        & 1,196                         & 12.2\%       \\
		\bottomrule
	\end{tabular}
\end{table}

\section{Sentence-BERT Embeddings} To generate embeddings for sentences, we applied the Sentence-BERT model \citep{reimers2019sentence}. We adopted the publicly available version of the model  \texttt{all-MiniLM-L6-v2} from the HuggingFace repository \footnote{\url{https://huggingface.co/sentence-transformers/all-MiniLM-L6-v2}}. Sentence-BERT produces a single embedding for a given unit of text. The resulting sentence embeddings serve as predictor variables in the regression model. 

\section{Llama~2 Embeddings} 
In this work, we extracted sentence embeddings from all three versions of the Llama~2 model (without additional training for dialog). Officially, they are named after how many parameters  (roughly) each of them contains: Llama-7B, Llama-13B, and Llama-70B. Embedding dimensions from Llama-7B, 13B, and 70B model are 4096, 5120, 8192, respectively. The first two can be fit into one A100 GPU, while for the largest 70B version, we applied 8-bit quantization \citep{dettmers2022llm} before loading it into GPU. For each token, we only kept the outputs from the very last attention layer. From embeddings, we took average as the pooling function (as oppose to use the embedding from the last non-padding token) over each dimension for all tokens in one sentence. This was demonstrated with better performance in the preliminary work.

\section{Additional Results} \label{appendix:additionalResults}
\begin{figure}[h]
	\centering
	\begin{subfigure}[t]{.8\textwidth}
			\includegraphics[width=\textwidth]{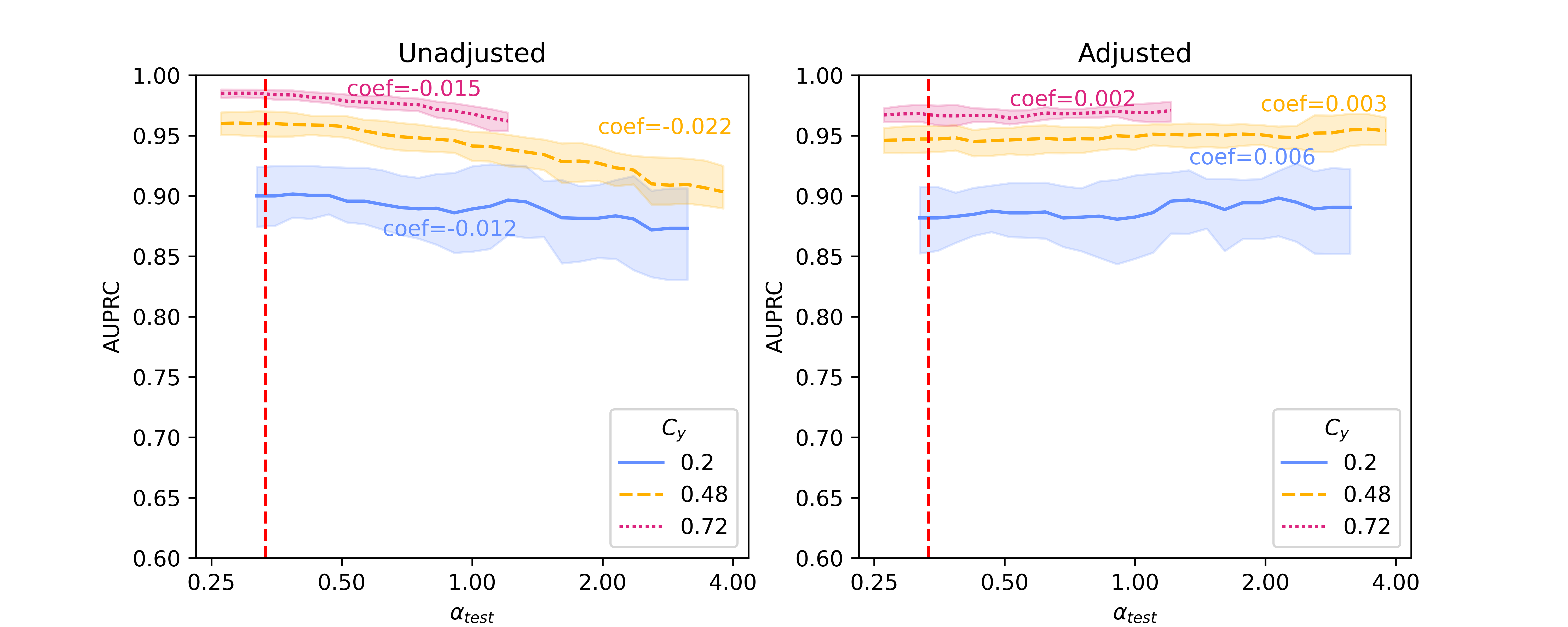}
			\subcaption{Llama 2-13B Average}
			\label{afig:SHAC_llama13}
		\end{subfigure}
		\begin{subfigure}[t]{.8\textwidth}
			\includegraphics[width=\textwidth]{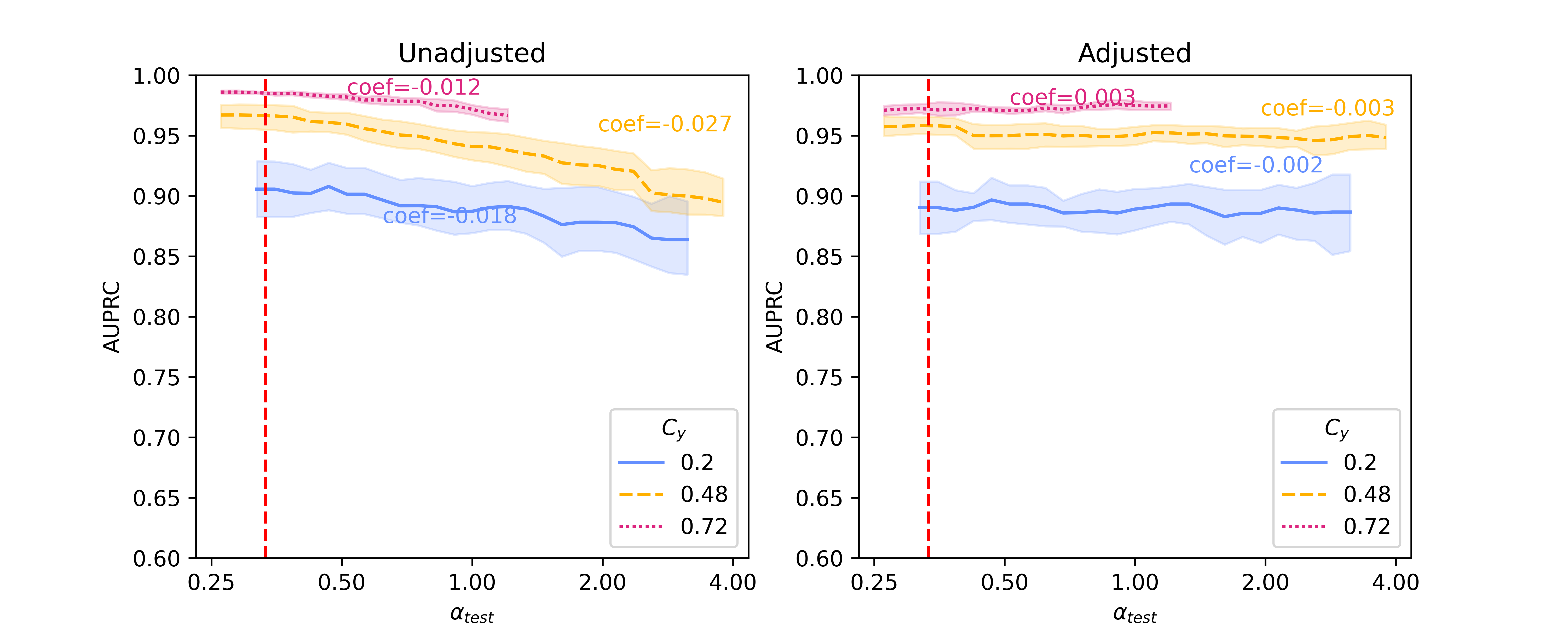}
			\subcaption{Llama 2-70B Average}
			\label{afig:SHAC_llama70q}
		\end{subfigure}
		\caption{Logistic regression and Backdoor Adjustment on SHAC dataset, based on (a) Embeddings generated from Llama~2-13B, then averaged; (b) Embeddings generated from quantized Llama~2-70B, then averaged. Red vertical line indicates $\alpha_{train}=0.33$, in which setting the model is trained.}
		\label{afig:SHACLR}
\end{figure}

\begin{figure}[h]
	\centering
	\begin{subfigure}[t]{.63\textwidth}
		\includegraphics[width=\textwidth]{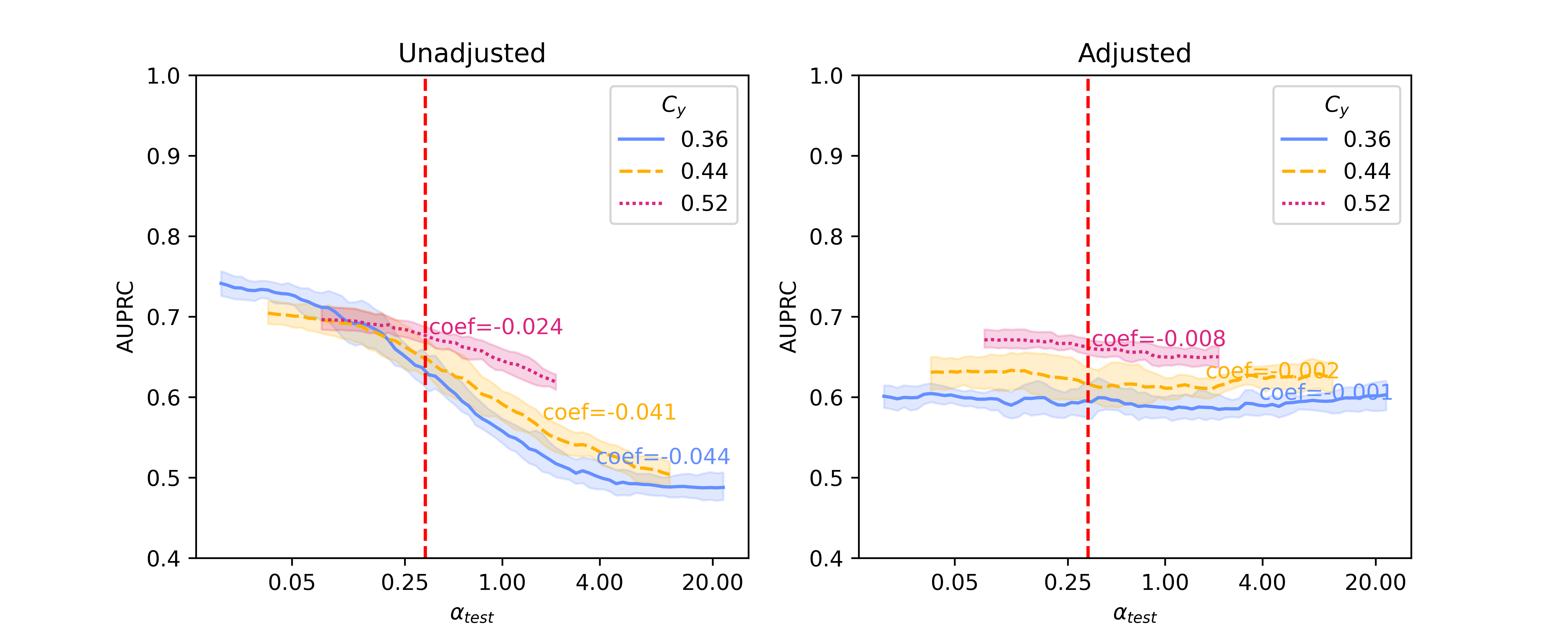}
		\subcaption{Binary Unigram}
		\label{afig:HS_binary}
	\end{subfigure}
	\begin{subfigure}[t]{.63\textwidth}
		\includegraphics[width=\textwidth]{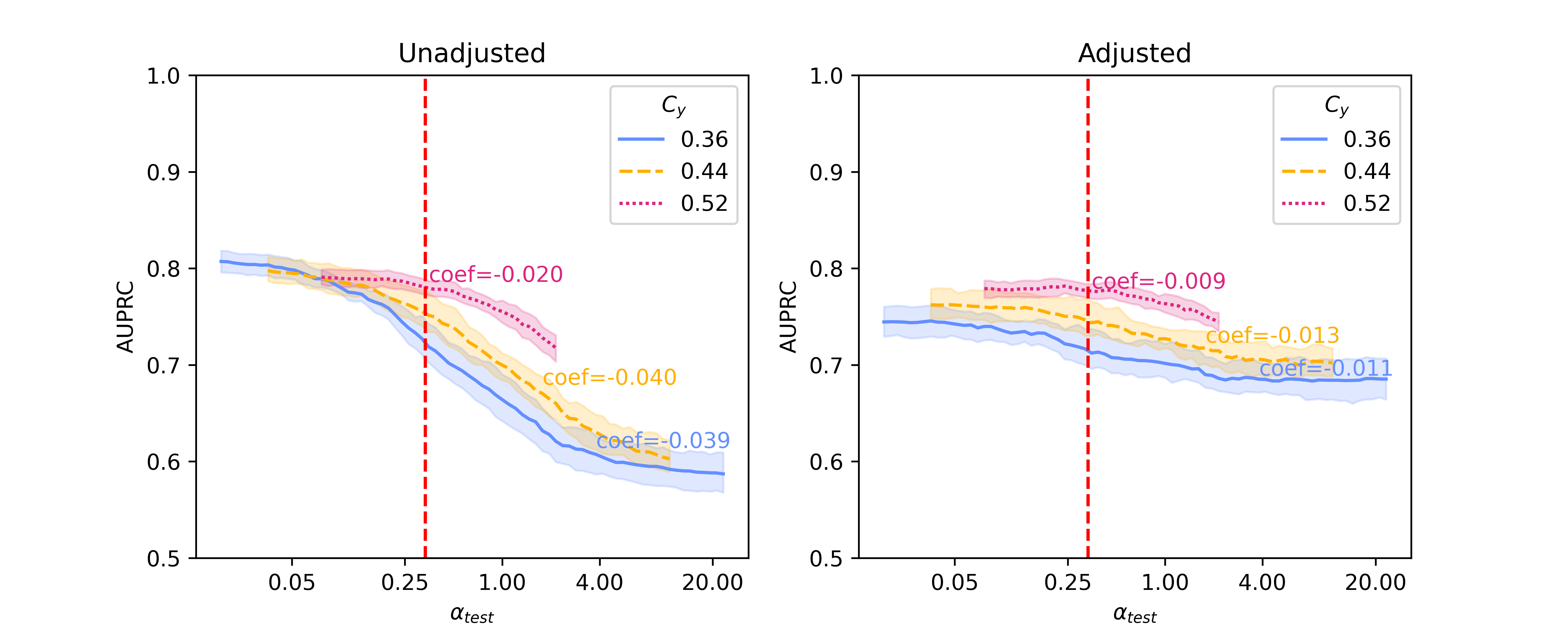}
		\subcaption{Sentence-BERT}
		\label{fig:HSLR_sentenceBERT}
	\end{subfigure}
	\begin{subfigure}[t]{.63\textwidth}
		\includegraphics[width=\textwidth]{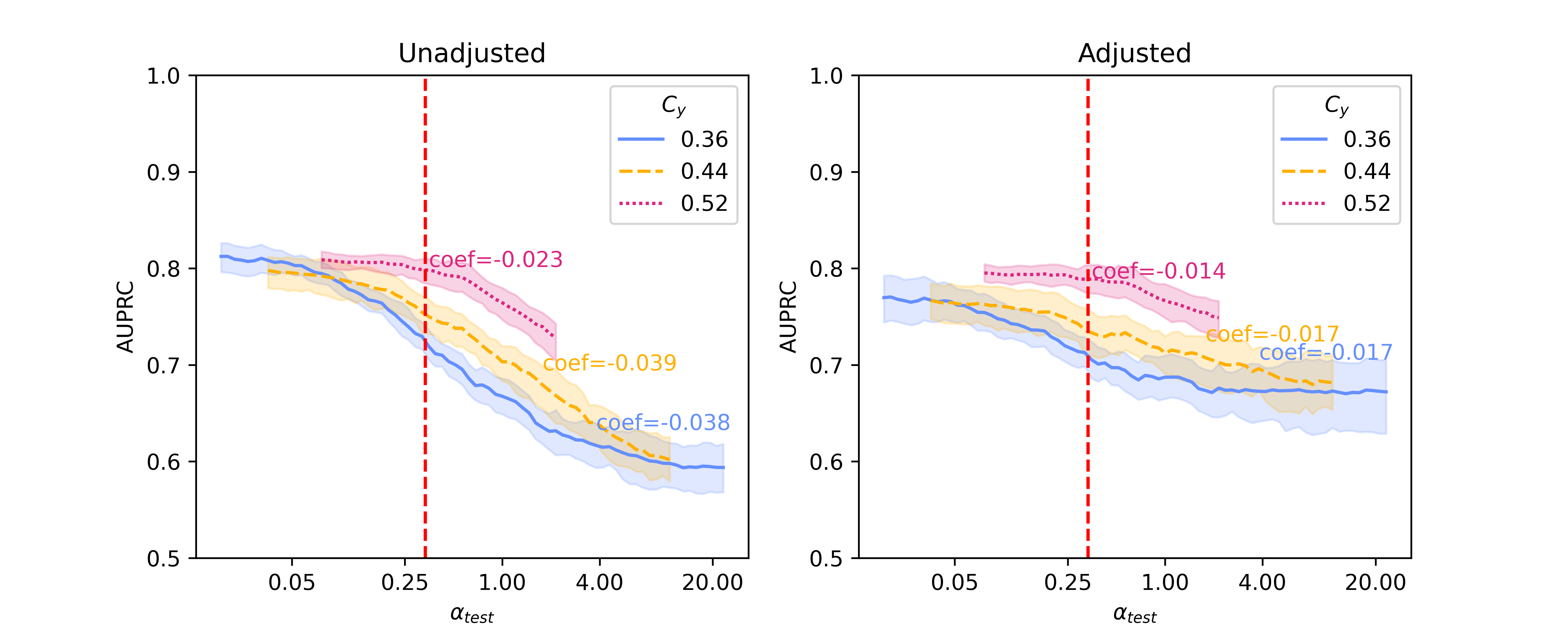}
		\subcaption{Llama 2-7B Average}
		\label{fig:HSLR_llama}
	\end{subfigure}
	\begin{subfigure}[t]{.63\textwidth}
		\includegraphics[width=\textwidth]{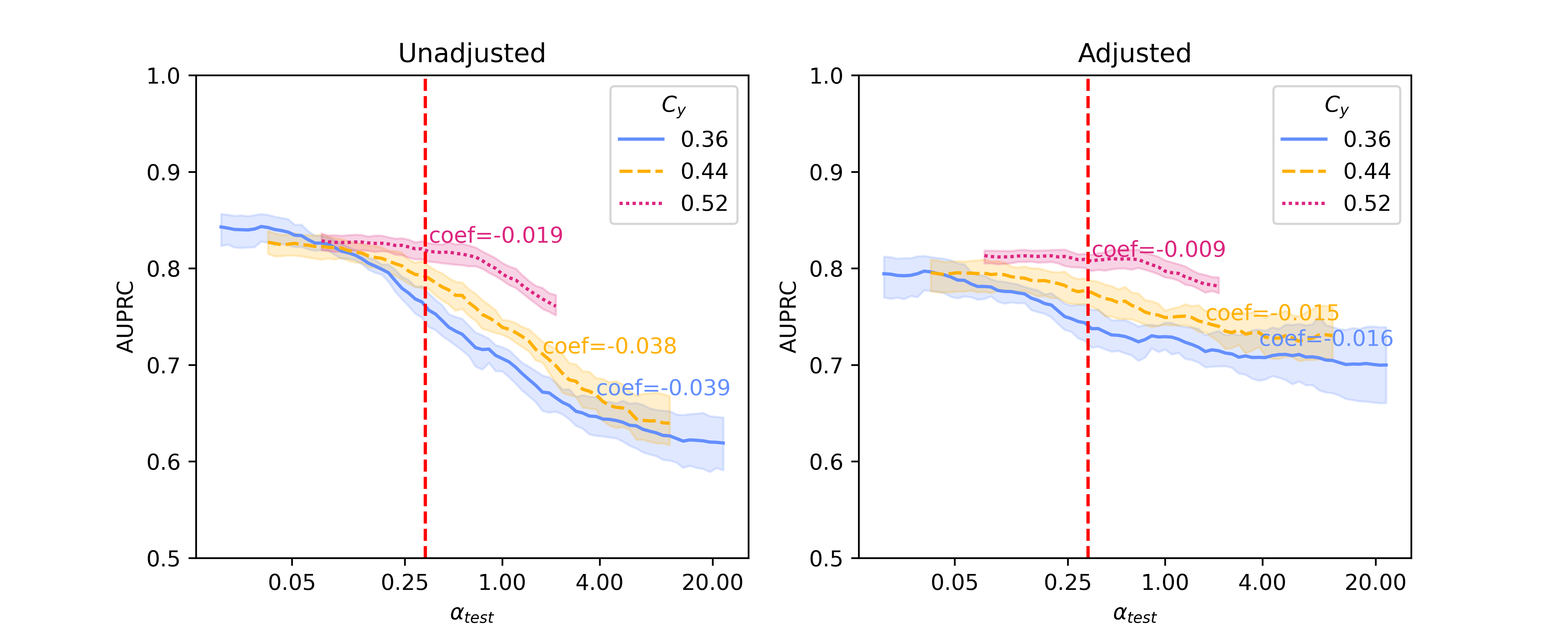}
		\subcaption{Llama 2-13B Average}
		\label{afig:HS_llama13}
	\end{subfigure}
	\begin{subfigure}[t]{.63\textwidth}
		\includegraphics[width=\textwidth]{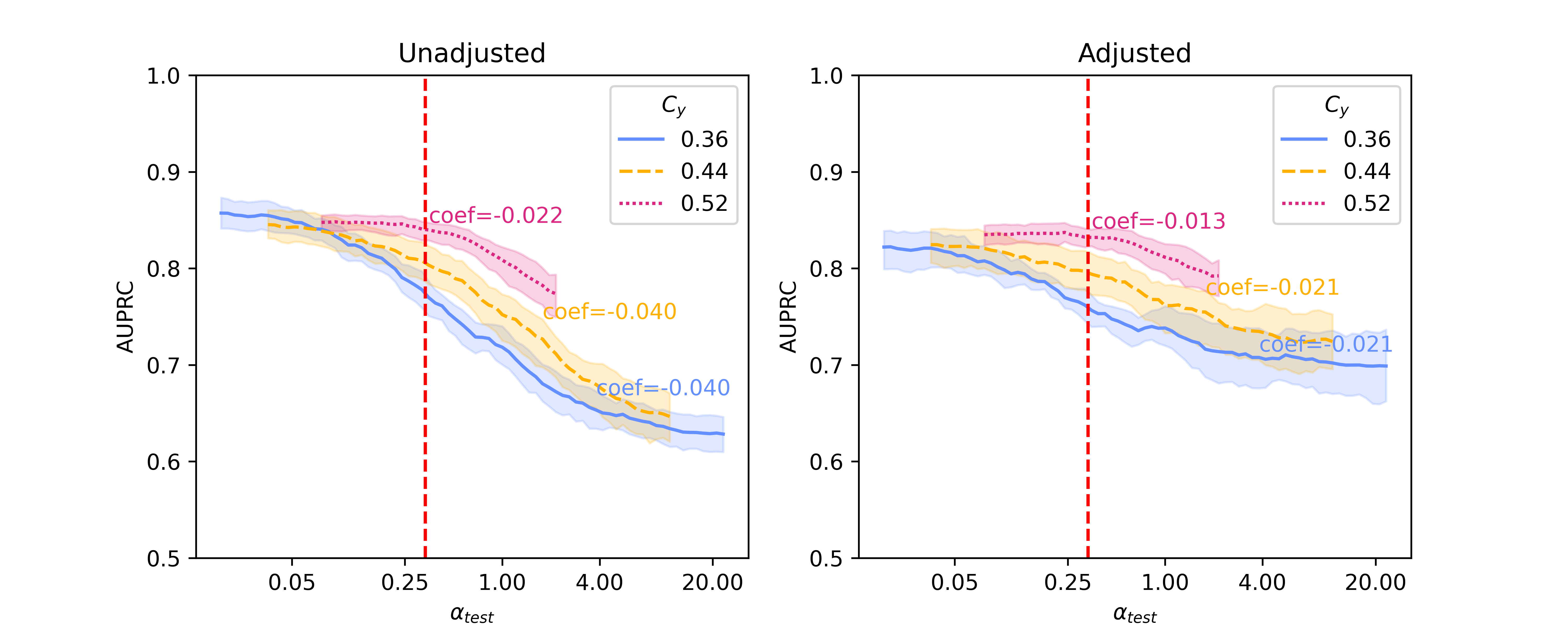}
		\subcaption{Llama 2-70B Average}
		\label{afig:HS_lama70q}
	\end{subfigure}
	\caption{Logistic regression and Backdoor Adjustment on the \textbf{Hate Speech Detection} dataset. (a) Binary Unigrams; (b) Embeddings generated from Sentence-BERT; (c) Embeddings generated from Llama~2-7B, then averaged; (d) Embeddings generated from Llama~2-7B, then averaged; (e )Embeddings generated from Llama~2-7B, then averaged. Red vertical line indicates $\alpha_{train}=0.33$, in which setting the model is trained.}
	\label{fig:HSLR}
	
\end{figure}

\begin{figure}[h]
	\centering
	\begin{subfigure}[t]{.8\textwidth}
		\includegraphics[width=\textwidth]{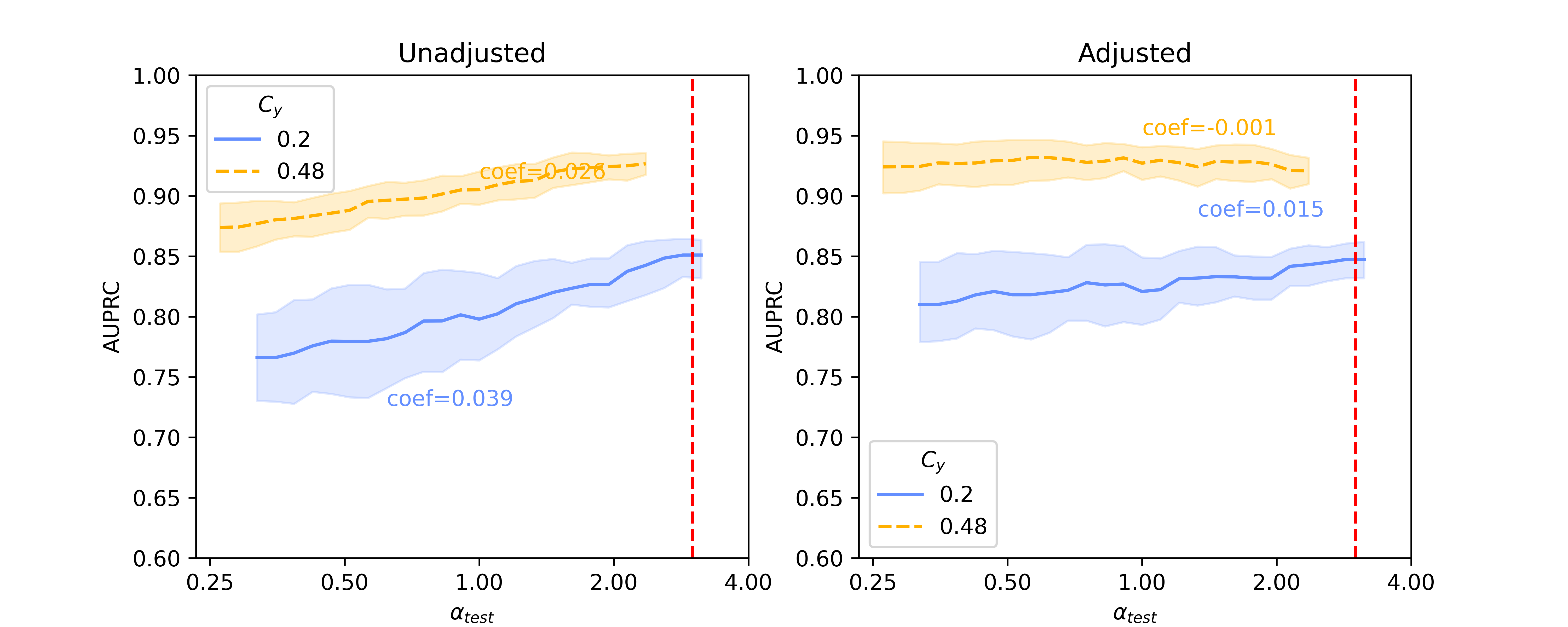}
		\subcaption{Binary Unigram}
		\label{afig:SHAC_binary_a3}
	\end{subfigure}
	\begin{subfigure}[t]{.8\textwidth}
		\includegraphics[width=\textwidth]{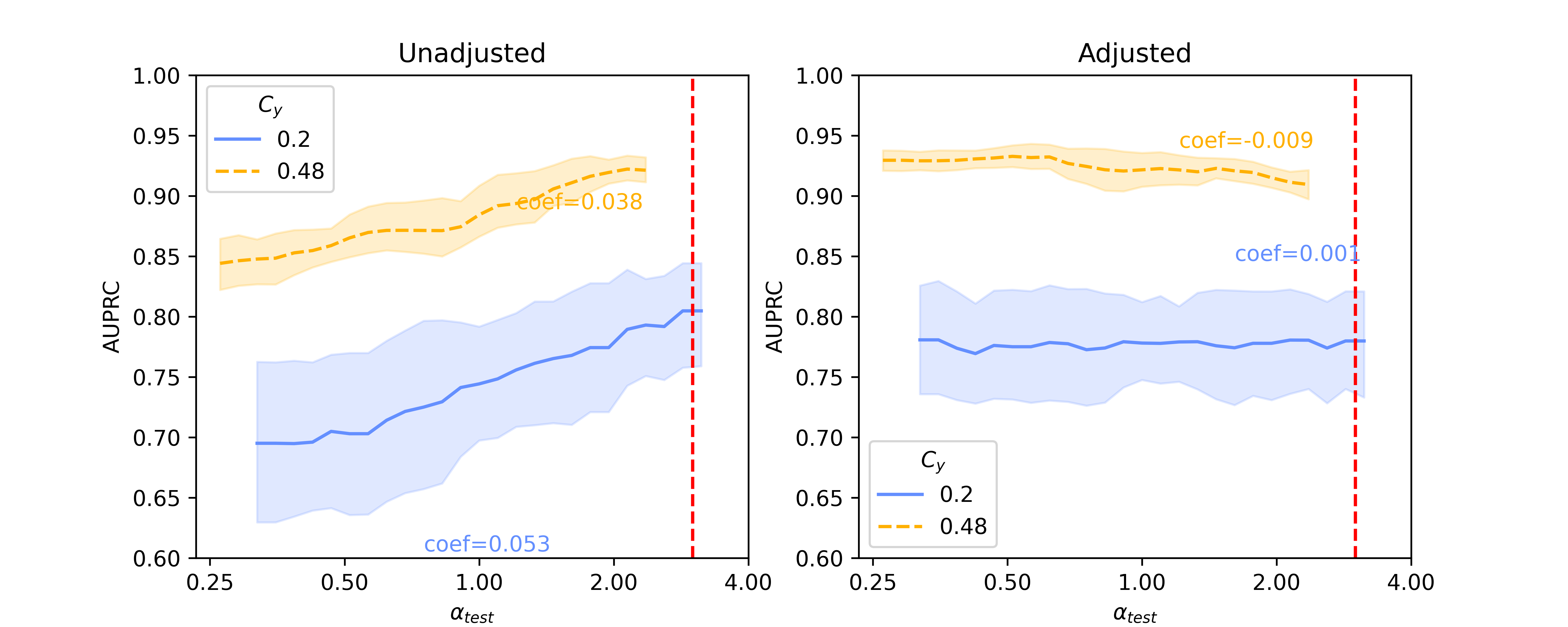}
		\subcaption{Sentence-BERT}
		\label{afig:SHACLR_sentenceBERT_a3}
	\end{subfigure}
	\begin{subfigure}[t]{.8\textwidth}
		\includegraphics[width=\textwidth]{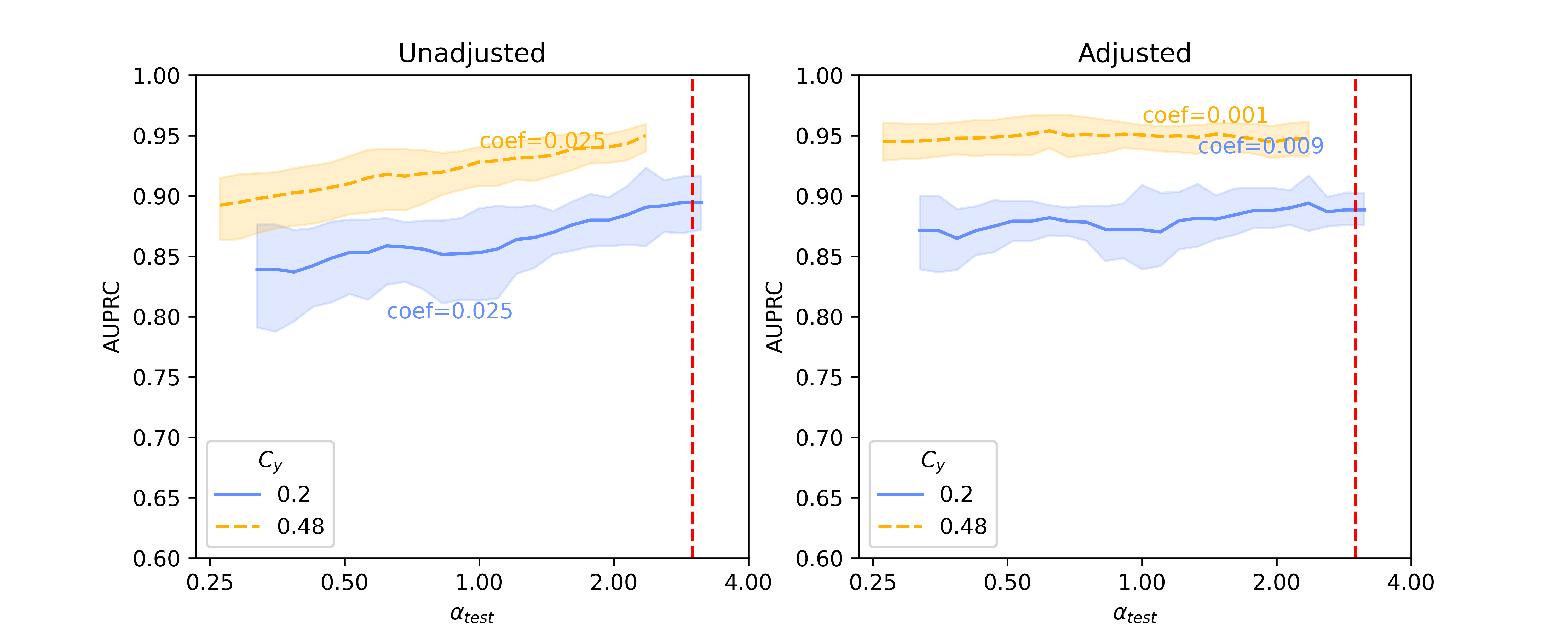}
		\subcaption{Llama 2-7B Average}
		\label{afig:SHAC_llama7b_a3}
	\end{subfigure}
	\begin{subfigure}[t]{.8\textwidth}
		\includegraphics[width=\textwidth]{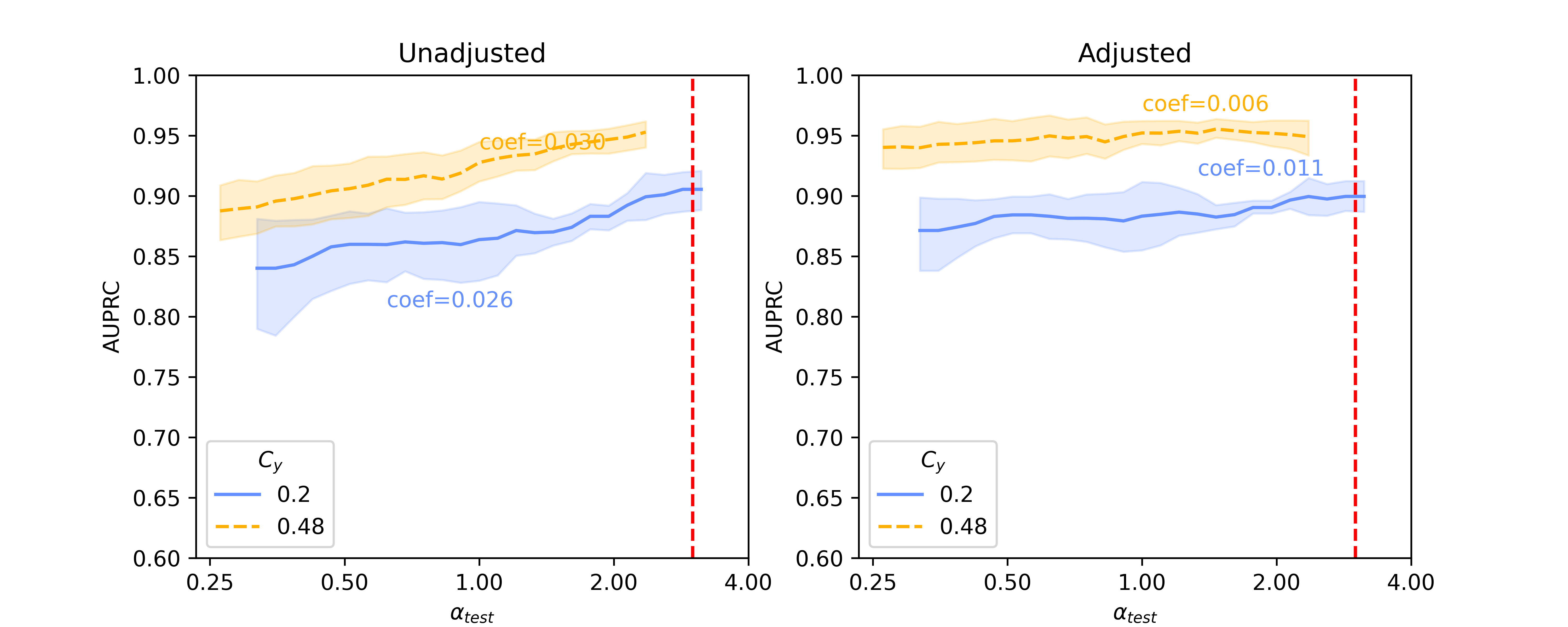}
		\subcaption{Llama 2-13B Average}
		\label{afig:SHAC_llama13b_a3}
	\end{subfigure}
	\caption{Logistic regression and Backdoor Adjustment on SHAC dataset, based on (a) Binary Unigrams; (b) Embeddings generated from Sentence-BERT; (c) Embeddings generated from Llama~2-7B, then averaged; (d) Embeddings generated from Llama~2-13B, then averaged. Red vertical line indicates $\alpha_{train}=\mathbf{3.0}$, in which setting the model is trained.}
	\label{afig:SHACLR_a3}
\end{figure}

\begin{figure}[h]
	\centering
	\begin{subfigure}[t]{.8\textwidth}
		\includegraphics[width=\textwidth]{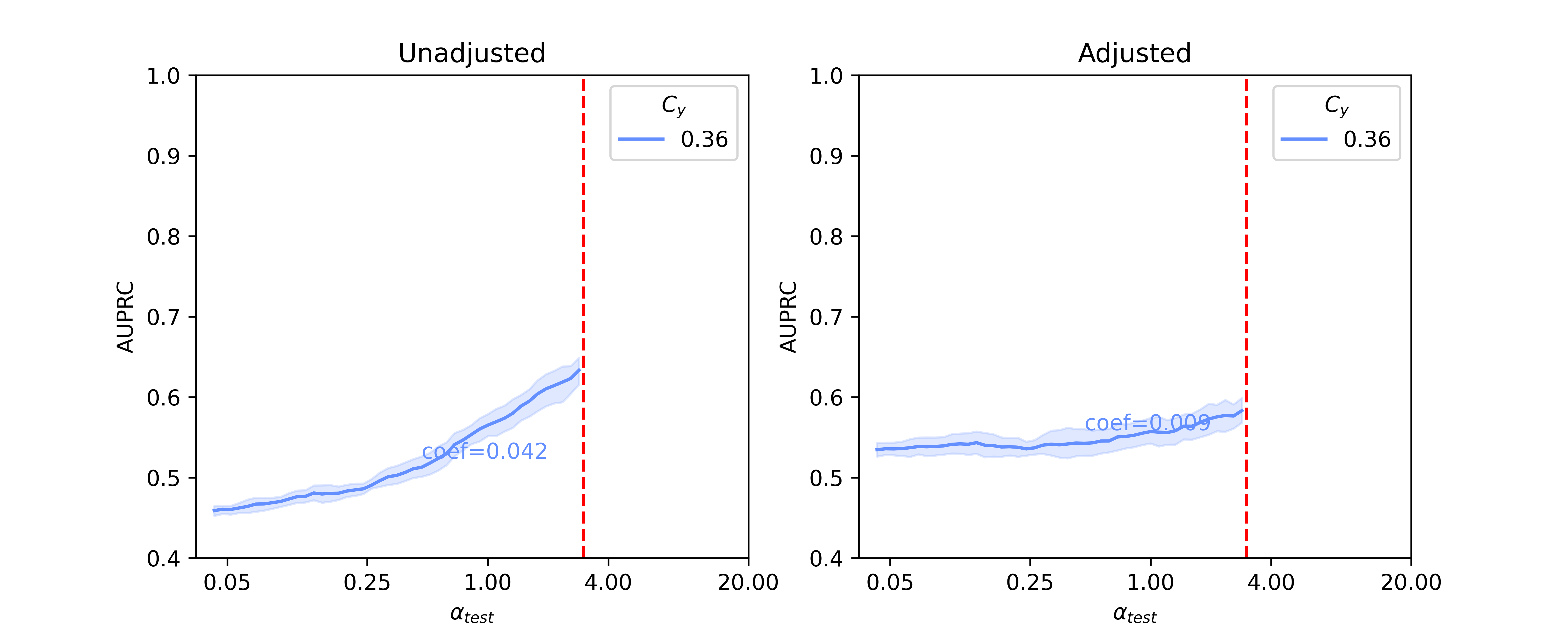}
		\subcaption{Binary Unigram}
		\label{afig:HS_binary_a3}
	\end{subfigure}
	\begin{subfigure}[t]{.8\textwidth}
		\includegraphics[width=\textwidth]{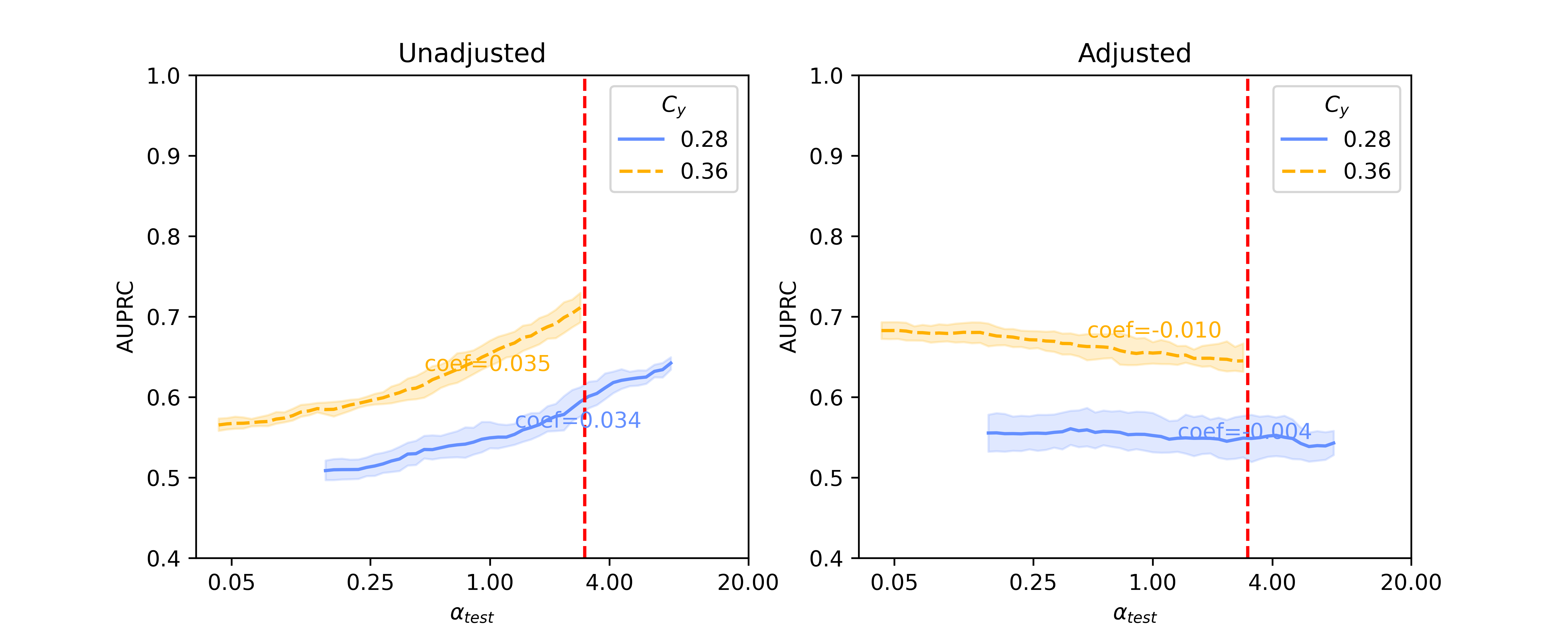}
		\subcaption{Sentence-BERT}
		\label{afig:HSLR_sentenceBERT_a3}
	\end{subfigure}
	\begin{subfigure}[t]{.8\textwidth}
		\includegraphics[width=\textwidth]{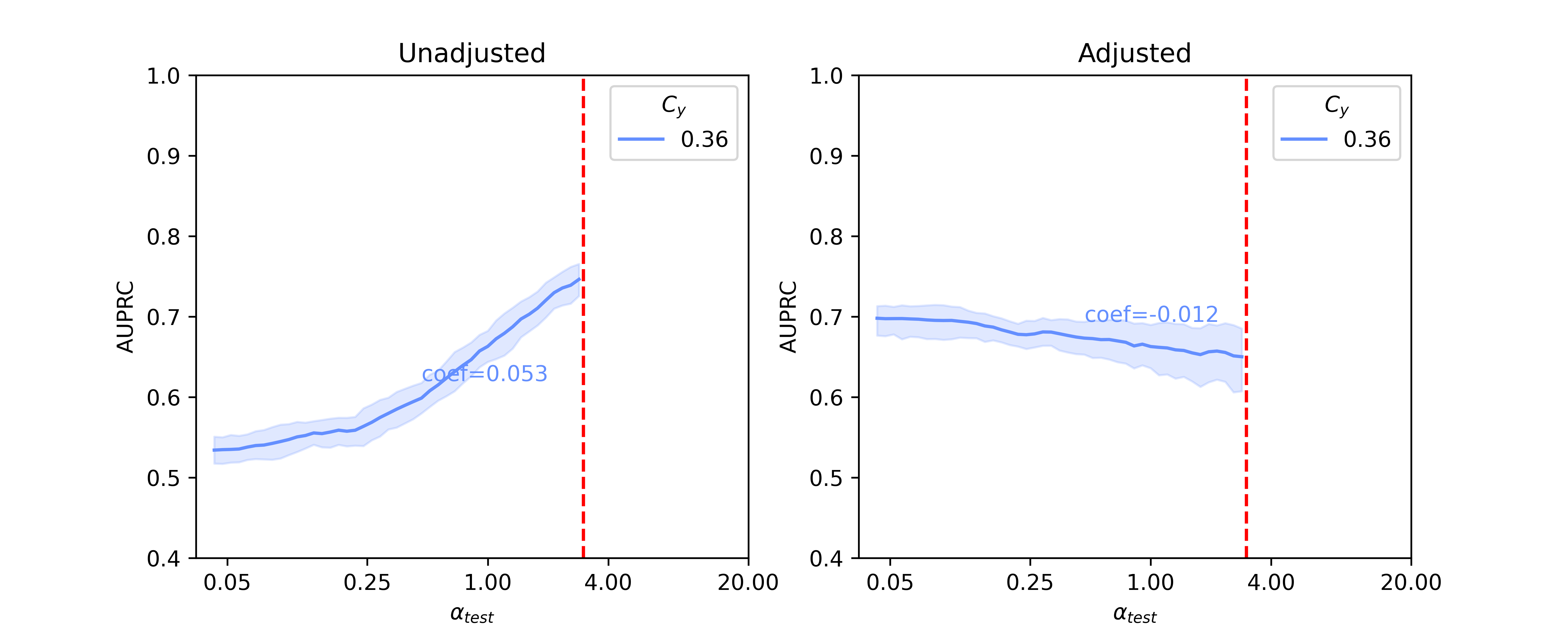}
		\subcaption{Llama 2-7B Average}
		\label{afig:HS_llama7b_a3}
	\end{subfigure}
	\begin{subfigure}[t]{.8\textwidth}
		\includegraphics[width=\textwidth]{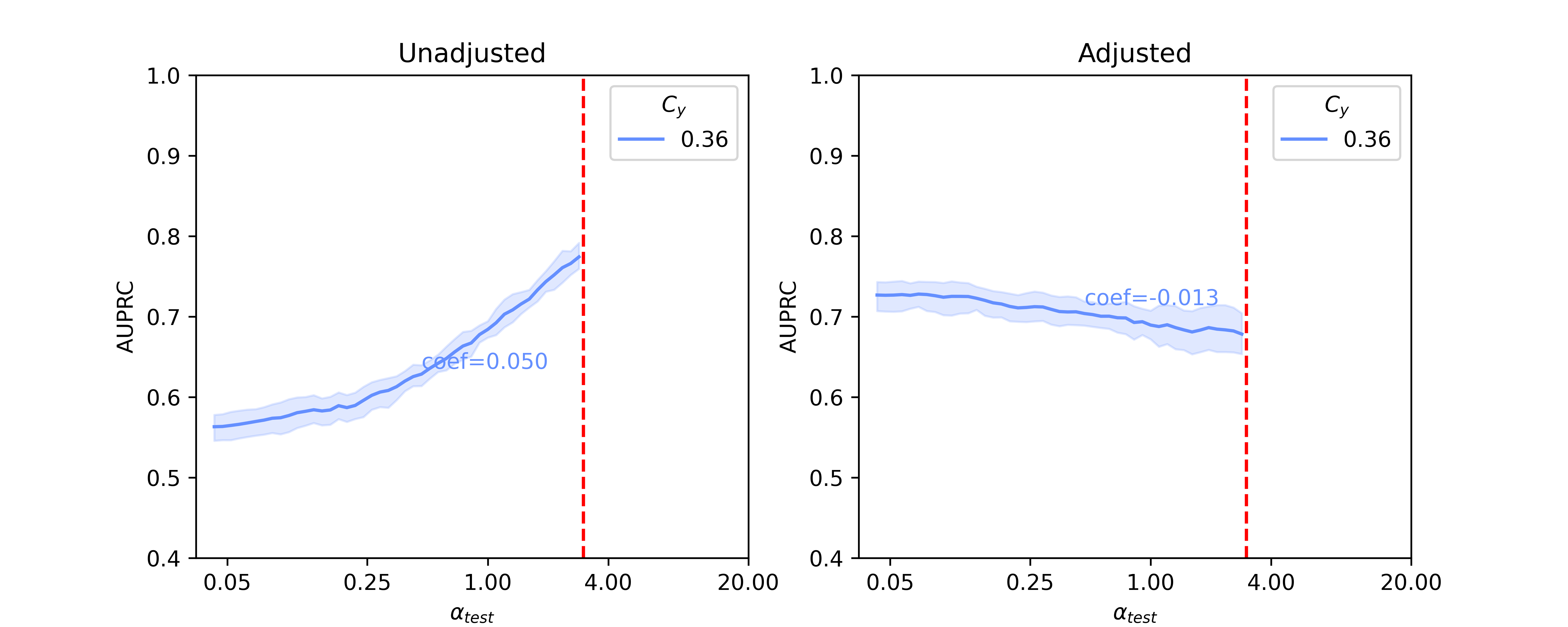}
		\subcaption{Llama 2-13B Average}
		\label{afig:HS_llama13b_a3}
	\end{subfigure}
	\caption{Logistic regression and Backdoor Adjustment on Hate Speech Detection dataset, based on (a) Binary Unigrams; (b) Embeddings generated from Sentence-BERT; (c) Embeddings generated from Llama~2-7B, then averaged; (d) Embeddings generated from Llama~2-13B, then averaged. Red vertical line indicates $\alpha_{train}=\mathbf{3.0}$, in which setting the model is trained.}
	\label{afig:HSLR_a3}
\end{figure}

In this section, we present baseline results on additional results on Llama~2-13B and quantized Llama~2-70B, for the SHAC dataset (Figure~\ref{afig:SHACLR}, coefficients presented in Table~\ref{atab:resultsSHACLR}) and full results for the Hate Speech Detection dataset (Figure~\ref{fig:HSLR}, coefficients presented in Table~\ref{atab:resultsHSLR}). Besides of results on $\alpha_{train}=0.33$, Figure~\ref{afig:SHACLR_a3} and Figure~\ref{afig:HSLR_a3} show additional results on $\alpha_{train}=3$, as the reciprocal according to Simulation Configurations in Appendix~\ref{appendix:simConfig}. It is noted that the value pair of 0.33 and 0.3 for $\alpha_{train}$  is randomly selected as an example and our interpretation holds true for other $\alpha_{test}$ values (not presented) with differences in degree of the effects. Different $\alpha_{test}$ ranges are shown for results on two datasets because empirical sampling may invalidate some of theoretical combinations.

For unadjusted models, embeddings from foundation models - Llama~2 models in particular - provide best innate robustness to confounding shifts. With respect to this robustness, Sentence-BERT lies at a similar level to the baseline binary unigrams. As for Llama~2 specifically, a larger model in size does not confer more robustness, as shown in our results on the two datasets where Llama~2-7B can produce lowest absolute coefficients in some scenarios ($C_y$ levels).

After applying Backdoor Adjustment to the logistic regression models, the absolute values of coefficients for all models drop, indicating better  model robustness to provenance-related distribution shifts. Among these, the baseline binary unigrams exhibit the most significant improvements after the adjustment, with the lowest absolute coefficients. However, for results on both datasets, the AUPRC measures from this baseline model are typically lower and deteriorate rapidly in comparison with those using foundation model representations. This performance drop is more significant with the Hate Speech dataset (Figure~\ref{fig:HSLR}).

When reviewing results for $\alpha_{train}=3$ in Figure~\ref{afig:SHACLR_a3} and Figure~\ref{afig:HSLR_a3}, we observe similar trends. Due to the sampling strategy, only some $C_y$ groups can be matched as in the case where $\alpha_{train}=0.33$. One apparent difference is that the slope of the lines is now mostly positive, as opposed to the negative slopes with $\alpha_{train}=0.33$. This indicates the training set can influence the model's baseline performance, in that when the testing set is more different from the training set (in terms of $\alpha$ values in our case), the performance usually drops. This matches the far right area when $\alpha_{train}=0.33$ and far left area for $\alpha_{train}=3$. 

\begin{table}[h]
	\caption{Coefficients for regression-based models on SHAC Dataset}
	\label{atab:resultsSHACLR}
	\centering
	\begin{tabular}{l|lccc}
		\toprule
		&                & \multicolumn{3}{c}{$C_y$}    \\
		&                & 0.20   & 0.48   & 0.72   \\
		\cmidrule(r){1-5}
		\multirow{5}{*}{Unadjusted LR} & Binary Unigram & -0.016 & -0.024 & -0.016 \\
		& Sentence-BERT  & -0.035 & -0.044 & -0.028 \\
		& Llama 2-7B     & -0.012 & -0.024 & -0.013 \\
		& Llama 2-13B    & -0.012 & -0.022 & -0.015 \\
		& Llama 2-70B    & -0.018 & -0.027 & -0.012 \\
		\midrule
		\multirow{5}{*}{Adjusted LR}   & Binary Unigram & 0.009  & -0.001 & 0.002  \\
		& Sentence-BERT  & -0.005 & -0.012 & -0.007 \\
		& Llama 2-7B     & 0.007  & 0.001  & 0.005  \\
		& Llama 2-13B    & 0.006  & 0.003  & 0.002  \\
		& Llama 2-70B    & -0.002 & -0.003 & 0.003  \\
		\bottomrule
	\end{tabular}
	\\
	{\scriptsize * Logistic Regression. $^\dagger$ Logistic regression with Backdoor Adjustment.}
\end{table}

\begin{table}[h]
	\caption{Coefficients for regression-based models on Hate Speech Dataset}
	\label{atab:resultsHSLR}
	\centering
	\begin{tabular}{l|lccc}
		\toprule
		&                & \multicolumn{3}{c}{$C_y$}    \\
		&                & 0.36   & 0.44   & 0.52   \\
		\cmidrule(r){1-5}
		\multirow{5}{*}{Unadjusted LR*} & Binary Unigram & -0.044 & -0.041 & -0.024 \\
		& Sentence-BERT  & -0.039 & -0.040 & -0.020 \\
		& Llama 2-7B     & -0.038 & -0.039 & -0.023 \\
		& Llama 2-13B    & -0.039 & -0.038 & -0.019 \\
		& Llama 2-70B    & -0.040 & -0.040 & -0.022 \\
		\midrule
		\multirow{5}{*}{Adjusted LR$^\dagger$}   & Binary Unigram & -0.001 & -0.002 & -0.008 \\
		& Sentence-BERT  & -0.011 & -0.013 & -0.009 \\
		& Llama 2-7B     & -0.017 & -0.017 & -0.014 \\
		& Llama 2-13B    & -0.016 & -0.015 & -0.009 \\
		& Llama 2-70B    & -0.021 & -0.021 & -0.013 \\
		\bottomrule
	\end{tabular}
	\\
	{\scriptsize * Logistic Regression. $^\dagger$ Logistic regression with Backdoor Adjustment.}
\end{table}

\clearpage

\section{Discussion} 
Overall, the empirical experiments on both datasets show consistent results. The logistic regression model, without any adjustment, is very sensitive to distribution shift, as evident in the higher absolute values of the coefficients of lines fit to the performance curves. However, different embedding methods show various degrees of robustness. When no adjustment is made, models trained on Llama~2 embeddings show best performance,  including the smaller 7B or 13B versions, especially when the dataset is small. 
As with the SHAC dataset, out-of-the-box usage of embeddings from foundation models (e.g., Llama~2-7B)  already shows some robustness to confounding shifts. However, without adjustment these representations do not provide significantly better robustness over the baseline unigram model (as in a larger Hate Speech Detection dataset). All these results suggest a need for adjustment to logistic regression models in the context of provenance-related confounding shift, irrespective of the choice of text representation technique.

Backdoor adjustment, when applied to the logistic regression, can significantly improve  robustness to these provenance-related shifts. This holds true for different embedding methods on the two datasets evaluated. Among them, binary unigrams benefits most from the adjustment, as shown in Figure~\ref{afig:SHAC_binary} when $C_y=0.48$, especially for the larger Hate Speech Detection dataset in Figure~\ref{afig:HS_binary}. However, this comes with a loss in performance of AUPRC, 
which in comparison is not significant for the SHAC dataset with very high baselines.

 One reason that Backdoor Adjustment works relatively well on unigrams  could be that foundation models generate highly contextual embeddings. In comparison, binary unigrams only store information at the unigram level from any corpus.  
 These two ways of representing text capture different information, which may further affect the distribution of representations. 
 Moreover, it should be noted that the embeddings from different models have different dimensions, which subsequently affect the size of a regression model ($\boldsymbol{\beta}$). As such, it could be argued that models trained on the resulting embeddings are not strictly comparable.  
 A potential way of controlling for this would be to apply dimension reduction methods, to enforce the constraint that all inputs to the regression model must have the same dimension. However, this will inevitably cause loss of information, with effects that are as yet unclear. There is another hyper-parameter, $v$, applied to the one-hot matrix for Backdoor Adjustment.  This serves as an added regulation term in regression models, and may have different effects with different numbers of parameters. In our work, $v$ was set to 10 for all experiments. The optimal $v$ value for models of different sizes remains to be determined.  Beyond our empirical experiments, formal testing of these assumptions is left to future work.

\section{Limitations}
In this work, we extended an evaluation framework for provenance-related confounding shift and examined one strategy for mitigating its effect on model stability. Several constraints were enforced during the development of the framework. They were originally set to isolate the effects of changes in site-specific class prevalence, but at the same time limit the scope of testing. Further work is also required to expand the framework to multiple classes and sources, other than binary cases discussed in this work. Moreover, it is left to future work to expand the framework to include other confounders, as in the case of the CivilComments dataset where a combination of 8 dimensions (e..g, gender, religion) collectively serve as the domain label \citep{borkan2019nuanced}.

The unprecedented size of foundation models (Llama 2 especially) limited our ability to utilize it fully in the context of available resources,  for example by preventing us from fine-tuning it end-to-end. The path we took in this work is a computationally lightweight approach and only serves as a first step into exploring Llama 2's potential. It remains to be determined whether such foundation models can be more robust when tuned using a parameter-efficient approach, or even fully fine-tuned.

Our primary focus has been on the application of foundation models to classification tasks in multi-institutional datasets within the biomedical domain, though several observations from our biomedical set align well with results from the general domain dataset. Further validation is required to determine how these findings generalize to different scenarios, including those with larger training samples and different model architectures, as well as whether different biomedical subdomains will favor different strategies.

\end{document}